\documentclass[conference]{IEEEtran}

\usepackage{cite}
\usepackage{amsmath,amssymb}
\usepackage{graphicx}
\usepackage{booktabs}
\usepackage{url}
\usepackage{xcolor}
\usepackage{enumitem}

\graphicspath{{figures/}}

\begin{document}

\title{Source Identification Is Not Fitness Testing:\\
Measuring the Limits of Synthetic-Data Attribution}

\author{
\IEEEauthorblockN{Joss Armstrong}
\IEEEauthorblockA{Ericsson, Athlone, Ireland\\
Email: joss.armstrong@ericsson.com}
}

\maketitle

\begin{abstract}
Repeated training on model-generated data can degrade later models. One
possible response is to use provenance when deciding which generated examples
to reuse. We test both how reliably that provenance can be recovered and
whether it helps identify better training data.

Using financial-risk text, we first identify the source of generated passages
and then repeat the test after rewriting them. Generator attribution is 98.7\%
accurate on the original passages but falls to 53.1\% after paraphrasing and
29.0\% after style rewriting. Generated-versus-human detection remains close
to perfect against the tested human comparison set.

We then compare two ways of selecting generated examples over three rounds of
generation and retraining. One uses source information. The other uses a score
from a separate reference model. The two rules select different examples, but
the planned comparison does not detect a stable difference in the degradation
of the resulting models.

The results show that identifying where data came from and identifying which
data are useful for training are separate problems.
The experiment therefore separates source identity, criterion-facing selection,
and recursive training outcome: neither the provenance score nor the tested
criterion-facing proxy is established as sufficient for future recursive
behaviour.
\end{abstract}


\section{Introduction}

Prior work has shown that repeatedly training models on model-generated data
can degrade later models \cite{shumailov2024collapse,alemohammad2024selfconsuming,kazdan2025collapse}.
The concern arises from the feedback loop itself. Outputs from one model can
become part of the data used to train a later model, so distortions introduced
in one round can be carried into the next. Provenance offers one possible way
to control what is reused. If generated material can be recognised before
retraining, its source can be considered when deciding which examples to
retain.

Provenance can answer two related questions. Is a passage human-written or
generated? If it is generated, which model produced it? We first test how well
these source questions can be answered and how the answers change after the
text is rewritten. We then ask a separate question: does source information
help choose examples that produce better models when they are reused for
training? To test that question, we compare a source-based selection rule with
a rule that scores each passage using a separate reference model. We then
repeat generation, selection, and retraining for three rounds and compare the
resulting models.

The source-identification experiments use financial-risk text generated by
several language models together with a human-written comparison corpus. We
measure generated-versus-human detection and generator attribution separately.
We then rewrite the generated passages in several ways to test how much source
information survives changes in wording and style. A further secondary test
examines whether the detector can recognise text from a source model that was
not used to train it.

The recursive experiment asks a different question. Every candidate example in
this part of the study is generated by the current model. One rule ranks the
examples using information from the source detector. A second rule ranks the
same pool using a score from a separate reference model. Each rule keeps half
of the generated examples. The retained data are used for retraining, and the
process is repeated for three generations. The resulting models are evaluated
on the same held-out human text.

This design keeps three questions separate: where the text came from, how the
candidate examples were ranked for selection, and what happened after training
on the selected data. The distinction matters because a source detector can be
accurate without establishing whether the detected source is relevant to the
later training outcome.
The recursive experiment is not a search for a sufficient statistic. Calling a
signal criterion-facing does not establish that it preserves every distinction
on which later recursive performance depends.

The problem is also relevant to communications systems. Existing standards
describe managed machine-learning pipelines, orchestration, and data-handling
functions for future networks \cite{itut2019y3172,itut2020y3174}. Generated
material may enter those workflows through model-assisted operations,
simulation, testing, documentation, or continual adaptation. The experiment
reported here uses financial-risk text as a controlled setting. It is not a
deployment experiment on telecommunications data.

The paper makes three contributions. First, it shows that generated-versus-
human detection and generator attribution are different source-identification
tasks and that exact generator identification degrades substantially after
paraphrasing and style rewriting. Second, it tests two further robustness
questions: whether detectors trained on known rewriting methods transfer to
other rewrites in the same family, and whether uncertain source assignments are
rejected rather than returned as one wrong label. Third, it compares two
selection rules in recursive training. The rules choose different examples,
but the planned comparison does not detect a stable difference in downstream
degradation. That result does not establish equivalence or show that provenance
is the better rule.

\section{Background and Related Work}

\subsection{Generated-text detection and source identification}

Prior work has generally treated generated-versus-human detection and
multi-generator attribution as different tasks. Uchendu et al.\
\cite{uchendu2020authorship} studied authorship attribution for neural text
generation, while Munir et al.\ \cite{munir2021looking} learned to identify the
source model of generated text. Both reported strong performance under
controlled conditions. Broader benchmarks such as RAID \cite{dugan2024raid}
and OpenTuringBench \cite{lacava2025openturingbench} have documented
sensitivity to domains, decoding choices, and rewriting. These results motivate treating generated-versus-human detection, identifying
which model produced the text, and recognising an unseen source as separate
measurements.

\subsection{Recursive training with generated data}

Shumailov et al.\ \cite{shumailov2024collapse} showed that models can degrade
when trained recursively on generated data. Alemohammad et al.\
\cite{alemohammad2024selfconsuming} studied related failure in self-consuming
generative models. Kazdan et al.\ \cite{kazdan2025collapse} showed that the
outcome depends strongly on the mixture of real and generated data and on the
training procedure. Gillman et al.\ \cite{gillman2024selfcorrecting} studied
self-correcting loops using preserved real data or correction functions. This
literature establishes that recursive reuse can matter. It does not establish
that provenance is the right basis for deciding which individual generated
examples should be retained.

\subsection{Rejecting uncertain source assignments}

We also test whether a source detector can decline to name a source when it is
uncertain. Split conformal prediction provides prediction sets with finite-
sample coverage guarantees under exchangeability
\cite{angelopoulos2023conformal}. A prediction set with several possible
sources can be treated as an abstention. Under distribution shift, however,
the coverage guarantee no longer applies in the same way, so a shifted example
may still receive a single wrong source label. We use conformal prediction to
distinguish abstention from these wrong singleton predictions.

\subsection{Three questions kept separate}

The paper keeps three questions distinct throughout:
\begin{enumerate}[leftmargin=*]
\item Where did the text come from? This includes generated-versus-human
detection, identification of the generating model, and recognition of a source
that was absent from detector training.
\item What score was used to decide which generated examples to retain for
training? The recursive experiment compares source-detector confidence with
perplexity under a frozen reference model.
\item What happened after training on the selected data? This is measured on a
separate frozen set of human-written financial-risk paragraphs.
\end{enumerate}

\section{Experimental Design}

\subsection{Corpus and source-model sets}

We began with a checksum-frozen corpus of financial-risk text generated by ten
models under a common prompt structure. An initial ten-generator attribution
sample used responses of at least 80 words, capped deterministically at 400
items per model and condition. The later robustness study used the seven
source models for which the full experimental sweep was complete, with 16{,}384
records per source: deepseek-v3.2, fastcontext-4b, gemma4-31b, gpt-oss-120b,
qwen2.5-14b-1m, qwen2.5-32b, and qwen3-coder-480b. Keeping these stages
separate is important because the initial ten-generator baseline and the later
seven-generator robustness experiments are not the same classification task.

\subsection{Human comparison corpus}

For generated-versus-human detection, we extracted 10-K Item~1A financial-risk
paragraphs filed in 2019 or earlier and length-matched them by token stratum.
For detector training, a deterministic cap of 20 paragraphs per ticker retained
879 paragraphs. The human material matches the financial-risk register but not
the exact generation task. The generated documents are summaries, while the
human documents are paragraph extracts. Every generated-versus-human result is
therefore scoped to this comparison corpus.

\subsection{Source-identification models}

We treated generated-versus-human detection, generator attribution, and
unseen-source rejection as separate tasks. The model set included linear
classifiers using word and character term frequency-inverse document frequency
(TF-IDF) features, combined word and character features, and a fine-tuned
DistilRoBERTa transformer classifier. The linear word model used unigram and
bigram TF-IDF features with sublinear scaling, a minimum document frequency of
3, and logistic regression. The character model used character $n$-gram
TF-IDF features with the same regularisation. Each classifier was trained and
evaluated independently.

\subsection{Rewriting conditions}

We rewrote generated texts in four ways intended to preserve their broad
content while changing surface form. Roundtrip translation translated the text
to a pivot language and back. Cross-family paraphrasing used a model from a
different family than the source model. Style transfer instructed the rewrite
model to match the register of human 10-K risk factors. Generic evasion asked
for a rewrite intended to make stylometric source detection more difficult
without giving access to the detector itself.

The threat model was grey-box. The rewriting model knew that stylometric source
identification was being used, but had no detector parameters, detector
training data, or query access, and produced one rewrite per item. For each
test condition, the rewriting model was excluded from source-detector training.
We first evaluated detectors trained only on the original source-identification
data. Secondary diagnostics then trained character-TFIDF models with examples
of the tested rewriting methods, measured transfer across those methods, and
held one rewrite type out at a time. All four transformations belong to the
same broad rewriting setup, so holding one out is not treated as evidence about
an entirely new class of transformation.

\subsection{Sources absent from detector training}

We also tested whether the detector could recognise that a text came from a
model absent from detector training. The first evaluation left one of the
seven reference source models out of training and then presented that model as
unknown. The second evaluation used five additional test-only generators, each
with 600 items, that were kept outside detector development. The first setting
trains on six known generators and the second trains on seven, so the resulting
areas under the receiver operating characteristic curve (AUROCs) are not
directly comparable.

\subsection{Abstention under uncertainty}

To test whether uncertain cases could be rejected rather than assigned a wrong
source, we used a split-conformal predictor with least-ambiguous-classification
scores \cite{angelopoulos2023conformal}. The predictor produced an eight-way prediction set at $\alpha=0.05$. We
treated an empty or multi-element set as an abstention and a singleton set as
a source assignment.
A wrong singleton rate is the fraction of all evaluated examples for which the
prediction set contains exactly one label and that label is wrong. We report
coverage, singleton rate, and wrong singleton rate alongside ordinary source-
identification accuracy.

\subsection{Recursive training experiment}
\label{sec:recursive_design}

The recursive study started every independent run from the same
Qwen2.5-3B-Instruct checkpoint and then carried out three generate-select-train
transitions, G0 to G3. We call each independent recursive run a lineage. At
each transition, the current checkpoint generated one financial-risk response
for each of 160 fixed prompts. The prompts used target windows of 120--160 and
100--140 words. Generation used temperature~1.0, top-$p$~1.0, top-$k$~0, no
truncation, and a preassigned lineage seed. Every candidate in this recursive
experiment was generated by the current model. The human corpus was used only
for downstream evaluation.

Each selection rule retained 80 of the 160 generated items and used the same
fine-tuning budget. The four rules were:
\begin{itemize}[leftmargin=*]
\item \emph{Provenance}: retain the 80 items with the lowest top-two posterior
margins from a frozen eight-way generator-plus-human source classifier. This
favours items that the detector ties least confidently to one known source
signature. In this recursive experiment it is not a generated-versus-human
filter because all candidate items are generated.
\item \emph{Perplexity}: retain the 80 items with the lowest perplexity under a
frozen Qwen2.5-1.5B-Instruct reference model. This is the arm named
``Criterion'' in the frozen experimental records. The score describes how
expected the text looks to one fixed reference model. It is not the downstream
outcome used to evaluate the retrained descendants.
\item \emph{Random}: retain a content-blind draw of 80 of 160, redrawn for each
cell using that cell's prime-number seed.
\item \emph{Fixed-draw control}: retain the same content-blind 80-of-160 draw,
using seed~7, across all lineages and generations. No arm retains the full
160-item pool.
\end{itemize}
The source classifier and reference model were fixed before the recursive run
and were never refitted.

Qwen2.5-3B-Instruct itself was not one of the seven source models used to train
the source classifier. Two Qwen-family source models, Qwen2.5-14B-1M and
Qwen2.5-32B, were present. The provenance score is therefore being applied to
outputs from a related but unseen model, which is important when interpreting
what its confidence means.

The provenance and perplexity rules used five independent prime-number seeds
(2, 3, 5, 7, 11). The random and fixed-draw controls used three (2, 3, 5).
Because G0 is the shared starting checkpoint and selection begins only for G1,
G2, and G3, these runs yield 48 evaluated rule-generation cells after G0. The
primary outcome was corpus perplexity on a frozen set of 200 pre-2020 human
financial-risk paragraphs under each resulting checkpoint. These paragraphs
were excluded from source-detector training, and no selection rule was
evaluated using the score that selected its own training data.

The primary comparison was fixed before the recursive run began. This was not a
public preregistration. For each lineage, we
fitted an ordinary least squares (OLS) slope of held-out perplexity against
generation. The primary test compared the mean slope under the perplexity rule
with the mean slope under the provenance rule using an exact two-sided
permutation over all $\binom{10}{5}=252$ assignments at $\alpha=0.05$. We
inverted the same test under a constant additive-shift model to obtain the
primary 95\% interval and used a Welch interval as a robustness check. A
separate simulation estimated what slope differences a five-versus-five design
could detect with useful power. Per-generation comparisons were secondary.
Retained-set overlap and score correlation were used only to check whether the
two rules were actually making different selections.

\section{Source Identification Before and After Rewriting}

Before rewriting, the source-identification classifiers were highly accurate.
In the initial ten-generator experiment, the mean pairwise area under the
receiver operating characteristic curve was 0.9995, with a minimum of 0.9811,
and the ten-way classifier achieved 98.15\% accuracy. The later seven-generator
rewriting study used a separate word-TFIDF classifier, which identified the
generating model with 98.7\% accuracy on the original passages.

Adding the human comparison corpus produced an eight-class classifier with
seven generator classes and one human-written class. Its overall accuracy was 98.81\%,
and every human paragraph in that test set was assigned to the human class.
In a separate generated-versus-human test, the minimum per-generator AUROC
exceeded 0.9999. These strong binary results apply only to the tested human
baseline, which matches the financial-risk register but not the exact generation
task.

\begin{figure}[!t]
\centering
\includegraphics[width=\columnwidth]{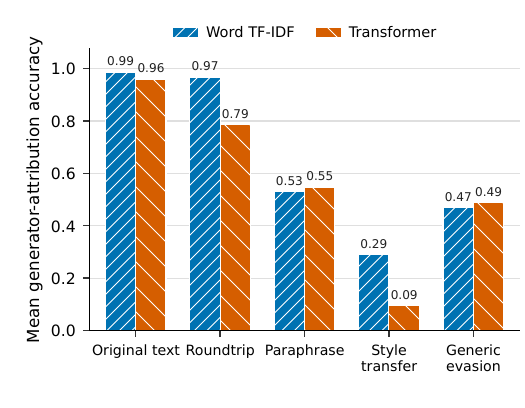}
\caption{Generator-identification accuracy before and after four forms of
rewriting. Values are means over seven source models. The word-TFIDF classifier
remains strong after roundtrip translation but falls after paraphrasing, style
transfer, and generic evasion. A separately trained transformer classifier performs the same generator-
identification task and falls to 0.094 after style transfer. Generated-versus-human detection is a separate task and is not
plotted.}
\label{fig:attribution_robustness}
\end{figure}

Rewriting changed the generator-identification result sharply
(Fig.~\ref{fig:attribution_robustness}). Averaged across the seven source models,
the word-TFIDF classifier fell from 98.7\% before rewriting to 96.7\% after
roundtrip translation, 53.1\% after paraphrasing, 29.0\% after style transfer,
and 46.9\% after generic evasion. A separately trained DistilRoBERTa classifier
performed the same generator-identification task. Its accuracy was 96.0\%
before rewriting and 9.4\% after style transfer. The failure under style
transfer was therefore not specific to the word-based classifier.
Table~\ref{tab:attrib-attack} gives the full comparison.

Generated-versus-human detection behaved differently from exact generator
identification. Against the tested human corpus, the binary TF-IDF AUROC
remained above 0.9999 under every rewrite. Rewriting could therefore remove
much of the evidence identifying which model generated a text while leaving the
evidence separating that text from this particular human baseline. This does
not establish universal generated-text detection.

\section{Training on Known Rewrites and Rejecting Uncertain Cases}

When examples of the tested rewriting methods were included in detector
training, a character-TFIDF classifier achieved 99.3\% to 99.8\% accuracy on
those represented transformations (Table~\ref{tab:attack-aware}). Holding one
of the four rewrite types out from this training gave mean accuracy of 95.7\%,
with the lowest value, 83.4\%, on held-out style transfer. These are useful
diagnostics of the tested rewrite family. They do not show that the detector
would remain robust to a new class of transformation.

Allowing the detector to abstain also did not remove every source error. At
$\alpha=0.05$, paraphrasing and generic evasion usually produced non-singleton
prediction sets. Their wrong singleton rates were 0.0027 and 0.0032. Style
transfer behaved differently. Its mean singleton rate was 0.5467 and its wrong
singleton rate was 0.4083, reaching 0.8567 for the worst source model
(Table~\ref{tab:conformal}). Under this shift, the detector frequently returned
one source label even when that label was wrong.

The practical conclusion is narrow. Performance on rewriting methods present
during detector development should not be treated as evidence for new forms of
rewriting. A rejection mechanism is also useful only when shifted cases are in
fact rejected rather than assigned one wrong source label.

\section{Recognising Sources Absent from Detector Training}

When one of the seven reference generators was excluded from detector training,
unseen-source AUROC ranged from 0.7700 to 0.9923, with mean 0.9261. The result
therefore depended substantially on which source was held out. For the five
additional generators kept entirely outside detector development, AUROC ranged
from 0.9714 to 0.9929, with mean 0.9822. Table~\ref{tab:openset} reports these
five values.

These results do not provide a guarantee for arbitrary future generators. They
show instead that recognising an unseen source is a different task from either
generated-versus-human detection or identifying one member of a known source
set.

\section{Recursive Training with Different Selection Rules}

At each recursive step, the current model generated 160 candidate texts. One
of four selection rules retained 80, the model was retrained on those examples,
and the resulting checkpoint was evaluated on the same held-out human corpus.
The main comparison was between the source-based rule and the reference-model
perplexity rule.

At G1, the perplexity rule produced lower held-out perplexity than the
provenance rule for all five shared seeds. This was a secondary descriptive
result rather than the primary comparison. The direction did not persist across
later generations, which is why the planned lineage-level trajectory comparison
is more informative than choosing one checkpoint.

The mean held-out-perplexity slope was 7.543 points per generation under the
perplexity rule and 7.367 under the provenance rule. The difference was
$+0.176$ points per generation. The exact two-sided permutation test did not
reject the null ($p=0.746$). The random-rule mean slope was 7.390, between the
two primary-arm means. Table~\ref{tab:lineage-slopes} lists the ten primary
lineage slopes and the three random-control slopes.

The lack of a slope separation was not because the two rules selected the same
examples. Their mean overlap was 36.0 of 80 retained examples at G1, 37.1 at G2, and
36.2 at G3. Fewer than half of the selected examples were shared between the
two rules at each generation. For two
independent uniformly random 80-of-160 selections, the expected overlap is 40,
which is used here only as a descriptive reference. The maximum absolute
correlation between source-detector confidence and reference-model perplexity
was 0.255 across the measured pools. The rules were making substantially
different selections.

The fixed-draw control also showed that the experiment could register recursive
change. Held-out perplexity increased from G0 to G3 in every fixed-draw lineage.
The two main rules nevertheless did not show a stable separation in their mean
recursive slopes. This result does not establish that they are equivalent. It
also does not establish that provenance is the better selection rule.

\section{What the Recursive Result Can and Cannot Show}

\subsection{Exact permutation test}

The primary comparison used an exact two-sided permutation over all
$\binom{10}{5}=252$ allocations of the ten observed primary-arm slopes into
two groups of five. The observed perplexity-minus-provenance difference was
$+0.176$ points per generation and the two-sided $p$-value was 0.746. Under the
null, the test requires exchangeability of the ten lineage slopes.

\subsection{Interval for the slope difference}

Inverting the exact permutation test under a constant additive-shift model gave
a 95\% interval of $[-0.794,+1.267]$ perplexity points per generation. A Welch
interval was $[-0.959,+1.311]$. These intervals still allow differences of
roughly one perplexity point per generation in either direction, so the
experiment does not support an equivalence claim.

\subsection{What differences could this design detect?}

A model-based sensitivity analysis used the observed pooled lineage standard
deviation of 0.716 in 200{,}000 balanced five-versus-five simulations under a
normal location-shift model. The exact two-sided permutation test was applied
in every replicate using seed~20260829. Estimated power was 0.157 for a true
0.5-point slope difference, 0.476 for 1.0 point, 0.799 for 1.48 points, and
0.964 for 2.0 points per generation (Table~\ref{tab:sensitivity}). The design
therefore had little ability to detect small differences and reached about 80\%
power only when the true difference approached 1.5 perplexity points per
generation. This calculation describes design sensitivity. It is not evidence
in favour of the null.

\section{Discussion}

\subsection{What source identification establishes}

The source experiments answer several different questions. Against the tested
human comparison set, generated-versus-human detection was close to perfect.
Identifying the exact generating model was also accurate on the original
passages, but that accuracy fell sharply after paraphrasing and style transfer.
Success on one source-identification task should therefore not be reported as
if it established success on the others.

\subsection{Robustness depends on what was tested}

When the detector was trained on examples of the tested rewriting methods, it
performed well on those methods. Holding out one rewrite type also gave strong
transfer within this small family, except for a larger drop on style transfer.
These results do not show how the detector would behave under a different kind
of rewriting. The abstention experiment adds a separate caution. Under style
transfer, many rewritten passages still received a single wrong source label
rather than an abstention.

\subsection{Different selections did not yield a stable recursive separation}

The provenance and perplexity rules made substantially different selections.
The recursive result therefore cannot be explained by the two rules choosing
the same data. The perplexity rule was included because it scores the text
itself rather than its source. That made it a plausible alternative selection
rule, but the experiment did not assume that the score contained everything
relevant to later training.

The first recursive generation illustrates why the longer trajectory matters.
The perplexity rule had lower held-out perplexity in all five shared-seed
comparisons at G1, but that direction did not persist. Across G0 to G3, the
planned lineage-slope comparison did not detect a difference between the two
rules. This does not show that provenance is superior, and the uncertainty is
too large to show that the two rules are equivalent.

\subsection{What the recursive result establishes}

Four conclusions are supported by the recursive experiment:
\begin{itemize}[leftmargin=*]
\item the provenance and perplexity rules selected substantially different
training data
\item the planned comparison did not detect different mean recursive slopes
\item the uncertainty interval still allows differences that could matter in a
larger or more sensitive experiment
\item neither source-detector confidence nor reference-model perplexity was
shown to contain all the information needed to predict the downstream effect of
training on the selected data
\end{itemize}

These conclusions apply to the two selection rules and the experimental design
tested here. They do not establish that either rule is a complete measure of
training value.

\subsection{Relevance to managed machine-learning systems}

Managed machine-learning pipelines in future communications systems already
include orchestration and data-handling functions
\cite{itut2019y3172,itut2020y3174}. Provenance can support audit, traceability,
and policy enforcement in those systems. Provenance alone does not establish whether retaining or removing a
particular generated example improves a later model. The financial-risk experiment illustrates that distinction. It
does not test network data or a deployed telecommunications system.

\subsection{The next question}

The unresolved problem is therefore not merely to replace provenance with
another fixed curation score. It is to determine what information a curation
rule must preserve for the declared downstream use, and whether that
information requirement remains valid after training changes the learner. A
signal can be criterion-facing without having been shown sufficient for future
recursive behaviour. This study identifies that boundary but does not
characterize the required recursive information interface.

A separate theoretical treatment
frames the same general issue in terms of task-relative information contracts,
where an information summary is judged by what it preserves for a named
downstream criterion \cite{armstrong2026tri}.

\section{Limitations}

\begin{itemize}[leftmargin=*]
\item \emph{One model scale.} The recursive experiment used
Qwen2.5-3B-Instruct. The result may not transfer to larger or smaller models.
\item \emph{One domain and task.} The experiment used financial-risk text. Its
communications relevance concerns the selection problem in managed training
pipelines, not performance on telecommunications data.
\item \emph{Three recursive updates.} G0 to G3 may not capture longer-horizon
effects.
\item \emph{Small lineage count.} Five lineages per primary rule provide
limited power for small differences. The simulation reached about 80\% power
only near a 1.48-point slope difference.
\item \emph{Four rewriting transformations.} All four tests are forms of text
rewriting. Holding one out does not approximate the full range of future
transformations.
\item \emph{Human comparison mismatch.} The human corpus matches the
financial-risk register but not the exact generation task. Every generated-
versus-human result is bounded by that comparison.
\item \emph{Recursive source model absent from detector training.}
Qwen2.5-3B-Instruct was not in the seven-model source-classifier roster. The
classifier included two related Qwen-family models, but its confidence on the
3B model should not be read as a calibrated probability of true source
membership.
\end{itemize}

\section{Conclusion}

Source identification worked well under the original test conditions. Exact
generator identification then weakened substantially after paraphrasing and
style rewriting, while generated-versus-human detection remained strong against
the tested human comparison set. These results show why the different source-
identification tasks should be evaluated separately.

The recursive experiment asked a different question. A source-based rule and a
reference-model-perplexity rule selected different training examples, but the
planned comparison did not detect a stable difference in the degradation of
the resulting models. The result does not establish that the rules are
equivalent, and it does not show that provenance is superior. Source information
can support provenance and filtering, but it does not by itself tell us which
generated examples will produce better models when reused for training.
Source-facing and criterion-facing selectors can be materially nonredundant
without either being established as sufficient for recursive curation.

\clearpage


\begin{figure}[!t]
\centering
\includegraphics[width=\columnwidth]{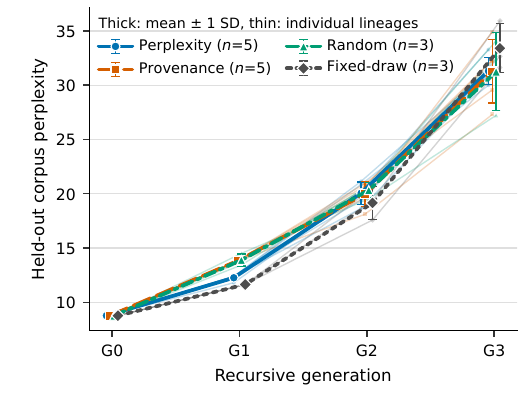}
\caption{Held-out corpus perplexity across recursive generations. Lower is
better. Thin lines show independent lineages and thick lines show rule means
with sample-standard-deviation error bars. The arm labelled ``Criterion'' in
the frozen experiment is the reference-model-perplexity rule described in the
text. Provenance and perplexity use five lineages each. Random and fixed-draw
use three each. All rules share G0. Mean slopes are 7.543 for perplexity and
7.367 for provenance, a difference of $+0.176$ perplexity points per generation.
The exact two-sided permutation test gives $p=0.746$ and the test-inversion
95\% interval is $[-0.794,+1.267]$. The interval does not establish
equivalence.}
\label{fig:recursive_trajectories}
\end{figure}


\begin{table}[!t]
\caption{Selected source-identification and robustness results}
\label{tab:robustness}
\centering
\small
\begin{tabular}{@{}p{0.40\columnwidth}p{0.29\columnwidth}r@{}}
\toprule
Evaluation & Measure & Result \\
\midrule
Seven-generator attribution before rewriting & Accuracy & 0.987 \\
Style transfer, word / transformer & Accuracy & 0.290 / 0.094 \\
Generated vs. human baseline & Minimum AUROC & $>0.9999$ \\
Unseen source, one roster model held out & AUROC range & 0.770--0.992 \\
Five sources outside detector development & AUROC range & 0.971--0.993 \\
Style transfer with conformal rejection & Wrong singleton rate & 0.408 \\
\bottomrule
\end{tabular}

\vspace{1mm}
\footnotesize Generated-versus-human detection, exact generator attribution,
and recognition of an unseen source are separate tasks and should not be read
as one combined accuracy measure.
\end{table}

\begin{table}[!t]
\caption{Generator-attribution accuracy before and after rewriting}
\label{tab:attrib-attack}
\centering
\small
\begin{tabular}{@{}lrr@{}}
\toprule
Condition & Word TF-IDF & Transformer \\
\midrule
Original text & 0.987 & 0.960 \\
Roundtrip translation & 0.967 & 0.787 \\
Paraphrase & 0.531 & 0.547 \\
Style transfer & 0.290 & 0.094 \\
Generic evasion & 0.469 & 0.489 \\
\bottomrule
\end{tabular}

\vspace{1mm}
\footnotesize Mean over seven source models. Generated-versus-human TF-IDF
AUROC remained above 0.9999 under every rewrite against the register-matched,
not task-matched, human comparison corpus.
\end{table}

\begin{table}[!t]
\caption{Character-TFIDF accuracy when rewrite types are represented or held out}
\label{tab:attack-aware}
\centering
\small
\begin{tabular}{@{}lrrr@{}}
\toprule
Rewrite & Original only & Included & Held out \\
\midrule
Roundtrip & 0.963 & 0.998 & 1.000 \\
Paraphrase & 0.510 & 0.998 & 0.998 \\
Style transfer & 0.490 & 0.993 & 0.834 \\
Generic evasion & 0.447 & 0.996 & 0.997 \\
\midrule
Mean & N/A & N/A & 0.957 \\
\bottomrule
\end{tabular}

\vspace{1mm}
\footnotesize ``Included'' means that examples of that rewrite type were
present during training. ``Held out'' means that the rewrite type was excluded
while the other tested rewrite types were represented. All four conditions are
forms of rewriting, so this is not evidence about a new class of
transformation.
\end{table}

\begin{table}[!t]
\caption{Conformal source rejection at $\alpha=0.05$, mean over seven source models}
\label{tab:conformal}
\centering
\small
\begin{tabular}{@{}lrrr@{}}
\toprule
Condition & Coverage & Singleton & Wrong singleton \\
\midrule
Original text & 0.942 & 0.947 & 0.004 \\
Roundtrip & 0.723 & 0.727 & 0.004 \\
Paraphrase & 0.055 & 0.058 & 0.003 \\
Style transfer & 0.138 & 0.547 & \textbf{0.408} \\
Generic evasion & 0.042 & 0.045 & 0.003 \\
\bottomrule
\end{tabular}

\vspace{1mm}
\footnotesize Wrong singleton rate is reported over all evaluated examples. A
wrong singleton is a one-label prediction set containing the wrong source.
Style transfer reaches 0.857 for the worst source model.
\end{table}

\begin{table}[!t]
\caption{Unseen-source AUROC by evaluation and generator}
\label{tab:openset}
\centering
\small
\begin{tabular}{@{}llr@{}}
\toprule
Tier & Generator & AUROC \\
\midrule
Roster holdout & \multicolumn{2}{l}{range 0.770--0.992, mean 0.926} \\
\midrule
Test-only & glm-5.2-fp8 (zai) & 0.971 \\
Test-only & kimi-k2.7-code (moonshot) & 0.982 \\
Test-only & qwen3-32b (qwen) & 0.993 \\
Test-only & deepseek-v4-flash (deepseek) & 0.987 \\
Test-only & Gemma2-27B-it (gemma) & 0.978 \\
\bottomrule
\end{tabular}

\vspace{1mm}
\footnotesize The first evaluation trains on six known sources. The second
trains on seven, so their AUROC values are not directly comparable.
\end{table}

\begin{table}[!t]
\caption{Per-lineage held-out perplexity slopes and primary comparison}
\label{tab:lineage-slopes}
\centering
\small
\begin{tabular}{@{}lr@{}}
\toprule
Lineage & Slope (perplexity points / generation) \\
\midrule
Perplexity s2 & 7.673 \\
Perplexity s3 & 7.372 \\
Perplexity s5 & 7.218 \\
Perplexity s7 & 7.258 \\
Perplexity s11 & 8.194 \\
\midrule
Provenance s2 & 5.991 \\
Provenance s3 & 6.955 \\
Provenance s5 & 8.424 \\
Provenance s7 & 7.754 \\
Provenance s11 & 7.711 \\
\midrule
Perplexity mean & 7.543 \\
Provenance mean & 7.367 \\
Difference & $+0.176$ \\
Exact $p$ (252 assignments) & 0.746 \\
95\% interval & $[-0.794,+1.267]$ \\
\midrule
Random s2 & 8.324 \\
Random s3 & 7.804 \\
Random s5 & 6.040 \\
\bottomrule
\end{tabular}

\vspace{1mm}
\footnotesize The perplexity rule is the arm named ``Criterion'' in the frozen
experimental records. Random mean slope is 7.390. Pooled standard deviation is
0.716.
\end{table}

\begin{table}[!t]
\caption{Retained-set overlap and score correlation}
\label{tab:overlap}
\centering
\small
\begin{tabular}{@{}lrr@{}}
\toprule
Diagnostic & Value & Random reference \\
\midrule
Overlap G1 (of 80) & 36.0 & 40.0 \\
Overlap G2 (of 80) & 37.1 & 40.0 \\
Overlap G3 (of 80) & 36.2 & 40.0 \\
Maximum absolute score correlation & 0.255 & N/A \\
\bottomrule
\end{tabular}

\vspace{1mm}
\footnotesize For two independent uniformly random 80-of-160 selections, the
expected overlap is 40. This value is a descriptive reference rather than an
inferential test. The correlation is between source-detector confidence and
reference-model perplexity across the measured pools.
\end{table}

\begin{table}[!t]
\caption{Estimated power by true slope difference}
\label{tab:sensitivity}
\centering
\small
\begin{tabular}{@{}rr@{}}
\toprule
True difference (perplexity points / generation) & Estimated power \\
\midrule
0.00 & 0.048 \\
0.50 & 0.157 \\
1.00 & 0.476 \\
1.48 & 0.799 \\
2.00 & 0.964 \\
\bottomrule
\end{tabular}

\vspace{1mm}
\footnotesize 200{,}000 balanced five-versus-five simulations under a normal
location-shift model with common standard deviation fixed to the observed
pooled value.
\end{table}


\clearpage

\bibliographystyle{IEEEtran}
\bibliography{references,references_strata_addition}

\end{document}